\documentclass[12pt,a4paper]{article}
\usepackage[dvips]{graphicx}

\begin{document}
\begin{center}
{\Large A COMPETITIVE COMPARISON OF DIFFERENT \\ 
TYPES  OF EVOLUTIONARY ALGORITHMS}\\ 
O. Hrstka$^{+}$ and A. Ku\v{c}erov\'{a}$^{+}$ and
M. Lep\v{s}$^{+}$\footnote[1]{%
Corresponding author. e-mail : {\tt leps@cml.fsv.cvut.cz}, fax:
+420-2-2431-077} 
and J. Zeman$^{\dag}$ \\
\vspace*{6mm}
$^{+}$Department of Structural Mechanics, Faculty of Civil
Engineering, Czech Technical University in Prague, Th\'{a}kurova 7,
166 29 Prague 6, Czech Republic\\
$^{\dag}$Department of Mechanics, Klokner Institute,
Czech Technical University in Prague, \v{S}ol\'{\i}nova 7,
160 00 Prague 6, Czech Republic
\end{center}
{\bf \large Abstract}
\\
This paper presents comparison of several stochastic optimization
algorithms developed by authors in their previous works for the
solution of some problems arising in Civil Engineering. The introduced 
optimization methods are: the integer augmented simulated annealing
(IASA), the real-coded augmented simulated annealing
(RASA)~\cite{Honza}, the differential evolution (DE) in its original
fashion developed by R.\,Storn and K.\,Price~\cite{Storn} and
simplified real-coded differential genetic algorithm
(SADE)~\cite{Sade}. Each of these methods was developed for some
specific optimization problem; namely the Chebychev trial polynomial
problem, the so called {\em type 0} function and two engineering
problems -- the reinforced concrete beam layout and the periodic unit
cell problem respectively. Detailed and extensive numerical tests
were performed to examine the stability and efficiency of proposed
algorithms. The results of our experiments suggest that the
performance and robustness of RASA, IASA and SADE methods are
comparable, while the DE algorithm performs slightly worse. This fact
together with a small number of internal parameters promotes the SADE
method as the most robust for practical use.


\section{Introduction}

Nowadays, optimization has become one of the most discussed topics of
engineering and applied research. Advantages coming from using
optimization tools in engineering design are obvious. They allow to
choose an optimal layout of a certain structure or a structural
component from the huge space of possible solutions based on a more
realistic physical model, while the traditional designing methods
usually rely only on some simple empirical formulas or guidelines
resulting in a feasible but not necessarily an (sub-)optimal
solution. Using optimization as a method of design can raise 
engineering job to a higher level, both in terms of efficiency and
reliability of obtained results. 

Typically, optimization methods arising in engineering design
problems are computationally demanding because they require
evaluation of a quite complicated objective function many times for
different potential solutions. Moreover, the objective function is
often multi-modal, non-smooth or even discontinuous, which means that
traditional, gradient-based optimization algorithms fail and {\em global
optimization techniques}, which generally need even a larger number of 
function calls, must be employed. Fortunately, the rapid development of 
computational technologies and hardware components allows us to treat
these problems within a reasonable time. 

As indicated previously, the optimization methods can be divided
generally into two groups: the gradient methods, that operate on a
single potential solution and look for some improvements in its
neighborhood, and global optimization techniques~--~represented here by
so called {\em evolutionary methods}~--~that maintain large sets
(populations) of potential solutions and apply some recombination and
selection operators on them. During the last decades, evolutionary
methods have received a considerable attraction and have experienced
a~rapid development. Good compendium of these methods can be
found for example in \cite{Evolutionary} and references therein. Main
paradigms are: {\em genetic algorithm} (binary or real coded), {\em
augmented simulated annealing} (binary or real coded), {\em evolution
strategy} and {\em differential evolution}. Each of these methods has
many possible improvements (see, e.g., \cite{Francouzi},\cite{Cinani}).   

Many researchers all over the world are united in an effort to
develop an evolutionary optimization method that is able to solve
reliably any problem submitted to it. In present time, there
is no such method available. Each method is able to outperform others
for certain type of optimization problem, but it extremely slows down
or even fails for another one. Moreover, many authors do not
introduce the reliable testing methodology for ranking their
methods. For example they introduce results of a single run of a given
method, which is rather questionable for the case of stochastic
algorithms.  Finally, the methods are often benchmarked on some test
functions, that even if presented as complicated, are continuous and
have few local extremes. 

This paper presents several optimization methods that were developed
and tested for different types of optimization tasks. The integer
augmented simulated annealing (IASA), derived from a binary version of
the algorithm \cite{Matej}, was developed to optimize a reinforced
concrete beam layout from the economic point of view. For solving the
problem of a periodic unit cell layout~\cite{Zeman:2001}, the
real-coded simulated annealing was applied. Differential evolution
arose to solve famous Chebychev polynomial
problem~\cite{Storn},\cite{Storn-WWW}. The last of the introduced
methods is the so-called SADE technology. It is a simplified
real-coded differential genetic algorithms that was developed as a
specific recombination of a genetic algorithm and a differential
evolution intended to solve problems on high-dimensional domains
represented by the {\em type 0} test function~\cite{Sade}. All these
methods may aspire to be a universal optimization algorithm. So, we
have conducted a detailed numerical tests of all these four
optimization methods for aforementioned optimization problems to
examine their behavior and performance.  

The paper is organized as follows. Section~2 provides brief description 
of each optimization task, while individual optimization algorithms are
discussed in Section~3. Numerical results appear in Section~4. Summary
on the performance of individual methods is presented in Section
5. For the sake of completeness, the parameter settings of algorithms
is shown in the Appendix. 

\section{Optimization tasks}

The optimization problems that are used as a set of test functions can be 
divided into two groups: the ``test suite'', containing ``artificial
functions'' and the ``engineering problems'', which collect more
(hopefully) practical optimization tasks. Specifically, these problems
are~: 
\begin{itemize}
\item Test suite:
\begin{itemize}
\item Chebychev trial polynomial problem,
\item {\em Type 0} benchmark trial function.
\end{itemize}
\item Engineering problems
\begin{itemize}
\item Reinforced concrete beam layout,
\item Periodic unit cell problem.
\end{itemize}
\end{itemize}
The following section provides description of selected functions in
more details.

\subsection{Chebychev problem}
%
%

\begin{figure}[!h]
\begin{center}
\includegraphics[width=14cm]{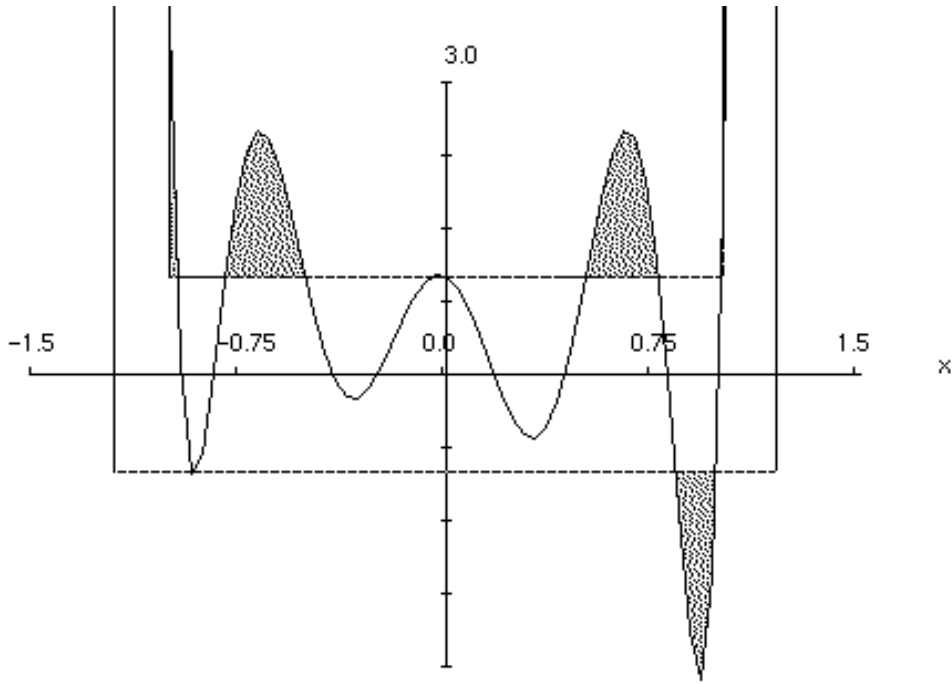}
\caption{A graph of a Chebychev polynomial ($n = 8$).}
\label{chebychev}
\end{center}
\end{figure}

Chebychev trial polynomial problem is one of the most famous
optimization problems. Our goal is to find such coefficients of a
polynomial constrained by the condition that the graph of the
polynomial can be fitted into a specified area (see
Fig.~\ref{chebychev}). 

Thus, the optimized values are the parameters $a_i$ of a polynomial
expression: 
\begin{equation}
f(x)=\sum_{i=0}^n a_i x^i,
\label{eq1}
\end{equation}
and the value of objective function is determined as a sum of
the areas, where the function graph exceeds a given boundary
(hatched areas in Fig.~\ref{chebychev}). 

\subsection{{\em Type 0} function}

\begin{figure}[!h]
\begin{center}
\includegraphics[width=14cm]{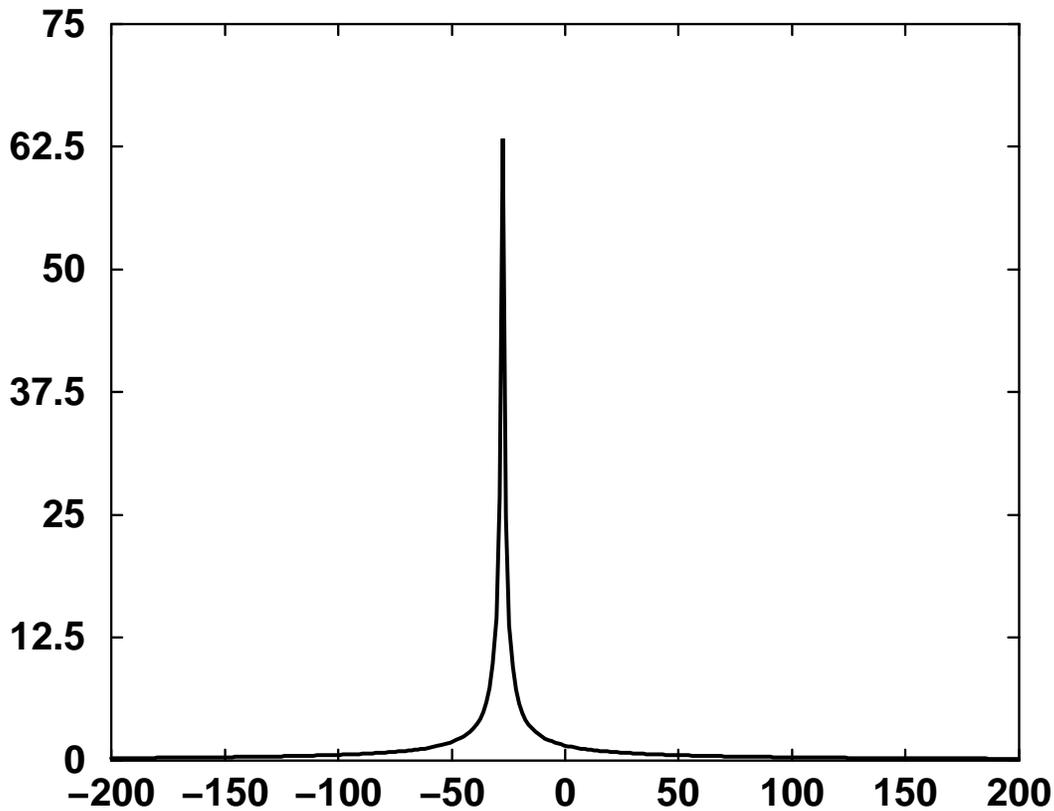}
\caption{An example of a {\em type 0} function.}
\label{Type0}
\end{center}
\end{figure}

This trial optimization problem was proposed by the first two authors
to examine the ability of the optimization method to find a single
extreme of a function with a~high number of parameters and growth
of computational complexity with the problem dimension. For this
reason, we used a function with a single extreme on the top of the
high and narrow peak:
\begin{equation}
f({\bf x})=y_0\left( \frac{\pi}{2}-\arctan\frac{\|{\bf x}-{\bf
x_0}\|}{r_0}\right),
\label{eq2}
\end{equation}
where ${\bf x}$ is a vector of unknown variables, ${\bf x_0}$ is 
the point of the global extreme (the top of the peak) and $y_0$ and
$r_0$ are parameters that influence the height or the width of the
peak, respectively. Example of such a function on one dimensional
domain is shown in Fig.~\ref{Type0}.

Although this example function has only a 
single extreme, to find it even with a~moderate precision is a
non-trivial task for several reasons. First, in the very
neighborhood of the extreme the function is so steep that    
even a futile change of the coordinates cause a large change of the
function value; in such a case it is very difficult for the algorithm
to determine what way leads to the top. Second, the peak is located
on a very narrow part of a domain and this disproportion increases
very quickly with the problem dimension.   

\subsection{Reinforced concrete beam layout}
\begin{figure}[!h]
\begin{center}
\includegraphics[width=14cm]{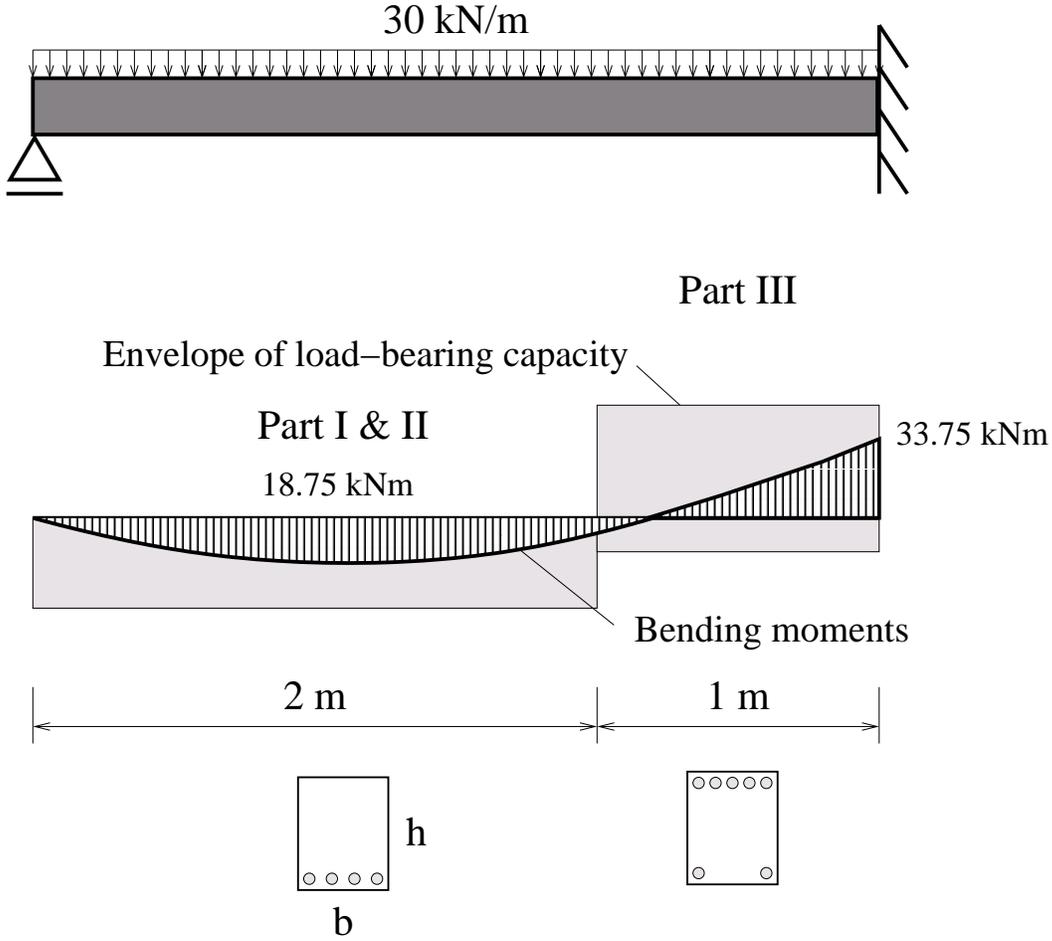}
\end{center}
\caption{A continuous beam subjected to an uniform loading.}
\label{f:matej:1}
\end{figure}

%
An effort to create an optimal design of a steel-reinforced concrete
structure is as old as the material itself. In present times an emphasis
is put on this problem due to widespread use of RC structures in Civil
Engineering. Frame structures are major part in this field with beams
playing an important role as one of the basic building block of this
construction system. An objective is to choose the best design from
all possible configurations that can create the requested
structure~--~in our case a~continuous beam (see Fig.~\ref{f:matej:1}).   

The total cost of the structure is used as a~comparison factor. An
advantage of the financial rating is its natural meaning to
non-experts and easiness of constraints implementation. In our
particular case, the objective function reads
\begin{equation}\label{eq:matej:1}
f({\bf X}) = V_cP_c + W_sP_s + \sum pf_i\;,
\end{equation}
where $V_c$ is the volume of concrete and $W_s$ is the weight of
steel; $P_c$ and $P_s$ are the price of concrete per unit volume and
steel per kilogram, respectively. From the mathematical point of view
the penalty function $pf_i$ is a~distance between a~solution and the 
feasible space, or equivalently, a price spent on the fulfillment of
the condition~$i$. Suppose that a variable $\Phi_i$ should not exceed
a certain allowable limit $\Phi_{i,max}$. Then, the penalty $p_{fi}$
assumes the form 
\begin{equation}
p_{fi} = \left\{ \begin{array}{ll}
0 & \mbox{if } \Phi_i \leq \Phi_{i,max}, \\
w_i \cdot \left( \Phi_i /\Phi_{i,max} \right)^2&  
  \mbox{otherwise},  
\end{array}\right.
\end{equation}
where $w_i$ is the weight of the $i$-th constraint.

The constraints in this procedure deal with
allowable strength and serviceability limits given by a chosen
standard --  in our case EUROCODE~2~(EC2) \cite{Eurocode}. An
interested reader can find implementation details for example in
\cite{Matej}. 

Consider a rectangular cross-section of a beam. There is the
width $b$ and the height $h$ to optimize. Other variables in
${\bf X}$ come from a model of a general RC beam which was 
presented in \cite{Honza} : the beam is divided to three
elements between supports with the same diameter of longitudinal  
reinforcement along the top surface and another one along
the bottom. The differences are only in numbers of steel reinforcement
bars in particular elements. The shear reinforcement is designed
alike. There are three shear-dimension parts - each of 
them with different spacing of stirrups but the same
diameter in the whole element. This partitioning reflects the
characteristic distribution of internal forces and moments in frame
structures, where the extremal values usually occur at three
points--at mid-span and two end joints. The novelty of our approach is
the assumption that length of parts may attain only the discrete values,
in our case corresponding to  0.025 m precision. The same principle is
used for the cross-section dimensions 
$b$ and $h$.  

\subsection{Periodic unit cell construction}
%

The motivation for this problem comes from the analysis of 
unidirectional fiber composite materials. Such materials consist of a
large number of fibers (which serve as a reinforcement) embedded in
the matrix phase. The goal is to determine the overall behavior of
such a material system provided that material properties of the matrix
and fibers are known. It turns out that for this prediction,
geometrical arrangement of fibers must be taken into account.  

\begin{figure}[!h]
\begin{center}
\includegraphics[angle = -90, width = 14cm]{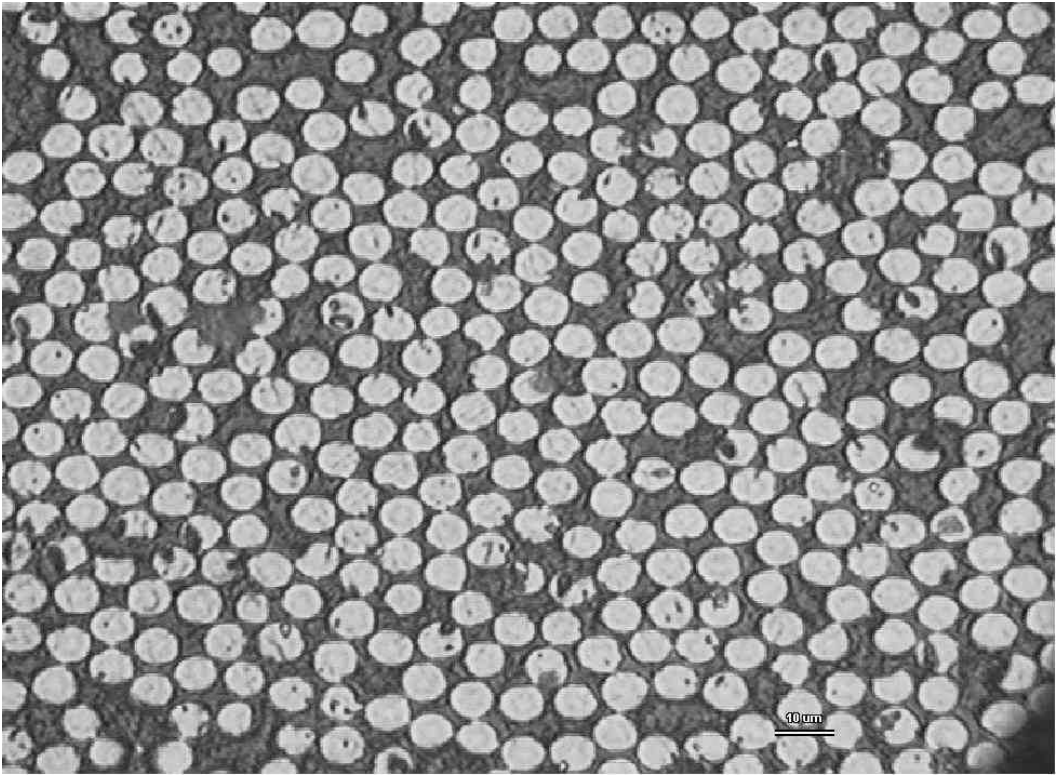}
\end{center}
\caption{An example of a microstructure of a unidirectional fiber  
composite.}
\label{f:puc:1}
\end{figure}

Unfortunately, the distribution of fibers in real composite materials
is quite complex (see Fig.~\ref{f:puc:1}). To avoid such an obstacle,
we attempt to replace a complicated microstructure with a large number of
fibers by a certain {\em periodic unit cell}, which resembles the
original material sample. More specifically, we 
describe the actual distribution of fibers by a suitable
microstructural function and then determine the parameters of the
periodic unit cell such that the difference between the function
related to the periodic unit cell and function related to the original
microstructure is minimized (for detailed discussion
see~\cite{Zeman:2001}).     

\begin{figure}[!h]
\begin{center}
\begin{tabular}{cc}
\includegraphics[width=10 cm]{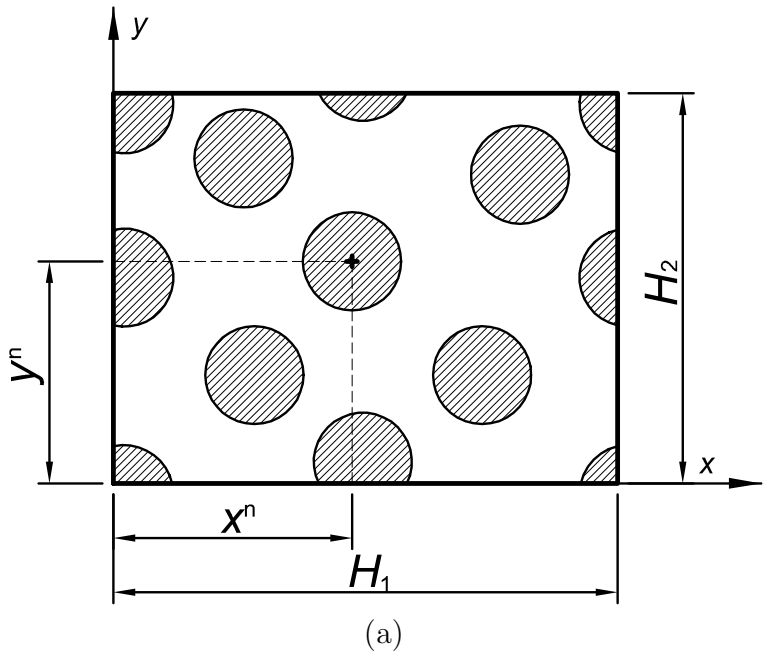} \\
(a)\\
\includegraphics[width=10 cm]{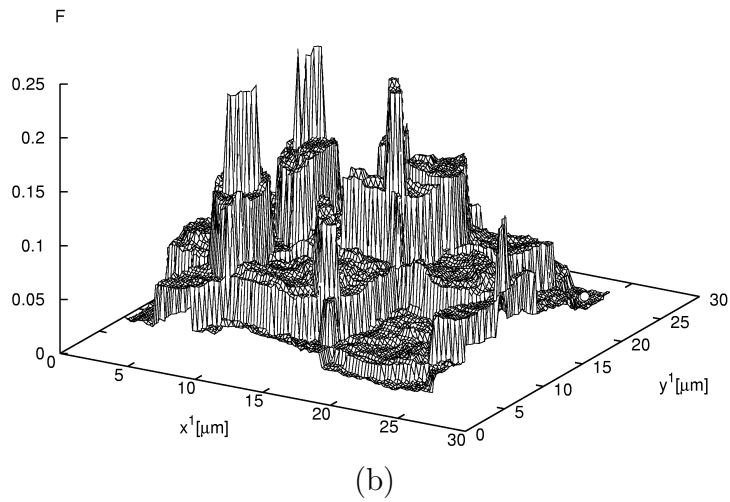} \\
(b)
\end{tabular}
\end{center}
\caption{(a) Geometry of a periodic unit cell, (b) An example of the
objective function.}   
\label{f:puc:2}
\end{figure}

The microstructural function used in the present approach is the {\em 
second order intensity function} $K(r)$, which gives the number of
further points expected to lie within a radial distance $r$ of an
arbitrary point divided by the number of particles  (fibers) per
unit area~(\cite{Ripley:1977}) 
\begin{equation}
\label{eq:puc:1}
K(r)=\frac{A}{N^2}\sum_{k=1}^{N} I_k(r),
\end{equation}
where $I_k(r)$ is the number of points within a circle with center at the
particle $k$ and radius $r$, $N$ is the total number of particles
(fibers) in the sample and $A$ is the sample area.
An objective function related to this descriptor can be defined as 
\begin{equation}
\label{eq:puc:2}
F( {\bf x}^N,H_1, H_2) = \sum_{i=1}^{N_m} \left(\frac{K_0(r_i)
- K(r_i)}{\pi r_i^2} \right)^2, 
\end{equation}
where vector ${\bf x}^N = \{x^1, y^1,\ldots,x^N, y^N\}$ stands for 
the position of particle centers of the periodic unit cell; 
$x^i$ and $y^i$ correspond to $x$ and $y$ coordinates of the
$i$-th particle, $H_1$ and $H_2$ are the dimensions of the unit cell
(see Fig.~(\ref{f:puc:2}a)), $K_0(r_i)$ is the value of $K$ function
corresponding to the original medium calculated in the point $r_i$ and
$N_m$ is the number of points, in which both functions are
evaluated. Throughout this study, we assume square periodic unit
cell ($H_1 = H_2$) and determine its dimensions in such a~way that the 
volume fraction of the fiber phase in the periodic unit cell is the
same as in the original micrograph. An example of the objective
function is shown in Fig.~\ref{f:puc:2}(b). 

\section{Applied methods}
%
During last few years, we have developed and tested several
evolutionary optimization methods that are based on both
binary/integer and real-valued representation of searched
variables. Each of them was primarily applied to one particular
optimization problem of the four introduced above. These methods are
(in order of appearance): 
\begin{itemize}
\item Differential evolution, developed by R.\,Storn and
K.\,Price~\cite{Storn},\cite{Storn-WWW} to solve the Chebychev trial
polynomial problem.

\item Simplified differential genetic algorithm, developed by authors for 
research on high-dimensional problems~\cite{Sade},\cite{Sade-WWW}.

\item Integer augmented simulated annealing - IASA (a combination of
integer coded genetic algorithm and simulated annealing); it was
primarily applied to the reinforced concrete beam layout optimization
problem. 

\item Real-coded augmented simulated annealing - RASA (a combination
of real-coded genetic algorithm by Michalewicz \cite{Michalewicz:1994}
and simulated annealing); it was developed for solving the periodic
unit cell problem. 
\end{itemize} 

\subsection{Differential evolution}
%
The differential evolution was invented as the solution method for the
Chebychev trial polynomial problem by R.\,Storn and K.\,Price in
1996. It operates directly on real valued chromosomes and uses the so
called differential operator, which works with real numbers in natural
manner and fulfills the same purpose as the cross-over operator in
the standard genetic algorithm. 

\begin{figure}[h]
\begin{center}
\includegraphics[width=14cm]{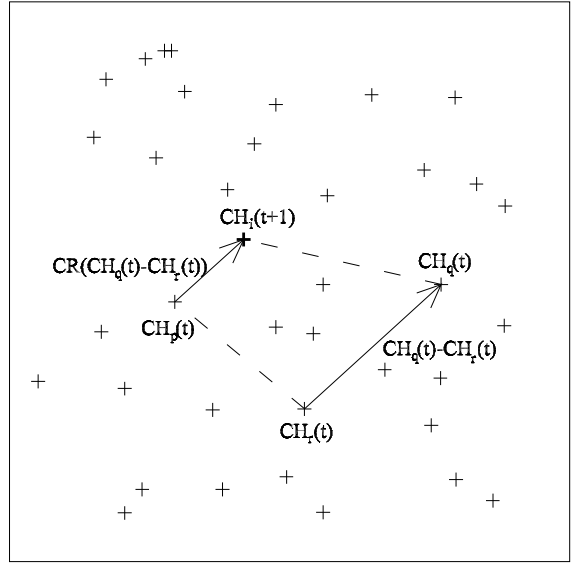}
\caption{The geometric meaning of a certain subtype of the
differential operator.} 
\label{operator}
\end{center}
\end{figure}

The differential operator has the sequential character: Let $CH_i(t)$
be the {\em i}-th chromosome of generation  {\em t} 
\begin{equation}
CH_i(t)=(ch_{i1}(t),ch_{i2}(t),...,ch_{in}(t)), 
\end{equation}
where $n$ is the chromosome length (which means the number of
variables of the fitness function in the real encoded case). Next, let
$\Lambda$ be a subset of $\{1,2,...,n\}$\footnote{The determination
of $\Lambda$ is influenced by the parameter called {\em
crossrate} ($CR$), see \cite{Storn}.}. Then for each $j\in\Lambda$
holds  
\begin{eqnarray}
ch_{ij}(t+1)=ch_{ij}(t) & + & F_1\left( ch_{pj}(t)-ch_{qj}(t)\right) 
\nonumber \\ 
& + & F_2\left( ch_{\mathrm{best}j}(t)-ch_{ij}(t)\right), 
\label{eq:doper1}
\end{eqnarray}
and for each $j\notin\Lambda$ we get
\begin{equation}
ch_{ij}(t+1)=ch_{ij}(t), 
\label{eq:doper2}
\end{equation}
where $ch_{pj}$ and $ch_{qj}$ are the {\em j}-th coordinates of two
randomly chosen chromosomes and $ch_{\mathrm{best}j}$ is the {\em
j}-th coordinate of the best chromosome in generation~{\em t}. $F_1$
and $F_2$ are then coefficients usually taken from interval
$(0,1)$. Fig. \ref{operator} shows the geometrical meaning of this
operator.
The method can be understood as a stand-alone evolutionary method or
it can be taken as a special case of the genetic algorithm. The
algorithmic scheme is similar to the genetic algorithms but  it is
much simpler: 
\begin{enumerate}

\item At the beginning an initial population is randomly created and
the fitness function value is assigned to each individual.  

\item For each chromosome in the population, its possible replacement
is created using the differential operator as described above.

\item Each chromosome in the population has to be compared with its
possible replacement and if an improvement occurs, it is replaced. 

\item Steps 2 and 3 are repeated until some stopping criterion is
reached. 

\end{enumerate}

As it could be seen, there are certain different features in contrary  
to the standard genetic algorithm, namely:
\begin{itemize}

\item the crossing-over is performed by applying the differential
operator~(\ref{eq:doper1},\ref{eq:doper2}), 

\item the selection operation like the roulette wheel, for example, is
not performed, the individuals that are going to be affected by the 
differential operator, are chosen purely randomly,

\item selection of individuals to survive is simplified to the
mentioned fashion: each chromosome has its possible replacement and if
an improvement occurs, it is replaced,

\item the mutation operator is not introduced as authors of DE claim
  that the differential operator is able to replace both mutation and
  uniform crossover known from basic GAs.

\end{itemize}

Further details together with the source codes of DE can be obtained
from web page~\cite{Storn-WWW}.

\subsection{Simplified atavistic differential evolution (SADE)}
%
This method was proposed as an adaptation of the differential
evolution in order to acquire an algorithm which will be able to solve
optimization problems on real domains with a high number of
variables. This algorithm combines features of the differential
evolution with classical genetic algorithms. It uses the differential
operator in the simplified form and an algorithmic scheme similar to
the standard genetic algorithm.  

The differential operator has been taken from the differential
evolution in a simplified version for the same purpose as the
cross-over is used in the standard genetic algorithm. This operator
has a sequential fashion: Let (again) $CH_i(t)$ be the {\em i}-th
chromosome in generation~{\em t}, 
\begin{equation}
CH_i(t)=(ch_{i1}(t),ch_{i2}(t),...,ch_{in}(t)), 
\end{equation}
where $n$ is the number of variables of the fitness function. Then, the
simplified differential operator can be written as 
\begin{equation}
ch_{ij}(t+1)=ch_{pj}(t)+CR\left( ch_{qj}(t)-ch_{rj}(t) \right),
\end{equation}
where $ch_{pj}$, $ch_{qj}$ and $ch_{rj}$ are the {\em j}-th
coordinates of three randomly chosen chromosomes and $CR$ is the so 
called {\em cross-rate\/}. Due to its simplicity this operator can be 
rewritten also in the vector form:  
\begin{equation}
CH_i(t+1)=CH_p(t)+CR(CH_q(t)-CH_r(t)).    
\end{equation}
Contrary to the differential evolution, this method uses the
algorithmic scheme very similar to the standard genetic algorithm:
\begin{enumerate}
\item As the first step, the initial population is generated randomly
and the fitness function value is assigned to all chromosomes in the 
population. 

\item Several new chromosomes are created using the mutation operators
- the mutation and the local mutation (number of them depends on a
value a of parameter called {\em radioactivity\/}, which gives the
mutation probability). 

\item Other new chromosomes are created using the simplified
differential operator as was described above; the whole amount of
chromosomes in the population doubles. 

\item The fitness function values are assigned to all newly created
chromosomes.  

\item The selection operator is applied to the double-sized
population, so the amount of individuals is decreased to its original
value. 

\item Steps 2-5 are repeated until some stopping criterion is
reached. 

\end{enumerate} 
Next, we introduce the description of the mentioned operators in
detail: 
\begin{description}

\item [Mutation:] If a certain chromosome $CH_i(t)$ is chosen to be
mutated, a new random chromosome $RP$ is generated and the mutated one
$CH_k(t+1)$ is computed using the following relation: 
\begin{equation}
CH_k(t+1)=CH_i(t)+MR(RP-CH_i(t)),
\end{equation}
where $MR$ is a parameter called {\em mutation-rate\/}.

\item [Local mutation:] If a certain chromosome is chosen to be locally mutated,
all its coordinates have to be altered by a random value from a
given (usually very small) range.  

\item [Selection:] This method uses modified tournament strategy to
reduce the population size: two chromosomes are randomly chosen,
compared and the worse of them is  rejected, so the population size is
decreased by $1$; this step is repeated until the population size
reaches its original size\footnote{Contrary to the classical
tournament strategy this approach can ensure that the best chromosome
will not be lost even if it was not chosen to any tournament.}.

\end{description}

Detailed description of the SADE method including source codes in
C/C++ and the tests documentation for the high-dimensional problems
can be obtained from the article \cite{Sade} and on the web-page
\cite{Sade-WWW}.

\subsection{Real-valued augmented simulated annealing (RASA)}
%
The augmented simulated annealing method is the combination of two 
stochastic optimization techniques -- genetic algorithm and simulated
annealing. It uses basic principles of genetic algorithms ( selection,
recombination by genetic operators ), but controls replacement of
parents by the Metropolis criterion (see Eq.~(\ref{eq:rasa1})). This
increases the robustness of the method, since we allow a worse child
to replace its parent and thus escape from local minima, which is in
contrary with DE methods described in Section 3.1.

The algorithmic scheme of the present implementation is summarized
as follows.

\begin{enumerate}
\item Randomly generate an initial population and assign a fitness to each
individual. Initial temperature is set to $T_0 = T_{max} = \verb!T_frac!
F_{avg}$ and minimal temperature is determined as $T_{min} =
\verb!T_frac_min! F_{avg}$ , where $F_{avg}$ is the average fitness
value  of the initial population.  

\item Select an appropriate operator. Each operator is assigned
a certain probability of selection.   

\item Select an appropriate number of individuals (according to the
operator) and generate possible replacements. To select individuals,
we apply {\em normalized geometric ranking} scheme (\cite{Houck:1995}):
The probability of selection of the $i$-th individual is given by
\begin{eqnarray}
p_i = q'(1-\mbox{\tt q})^{r-1}, && q'=
\frac{\mbox{\tt q}}{1-(1-\mbox{\tt q})^{\tt pop\_size}},
\end{eqnarray}
where \verb!q! is the probability of selecting the best individual in the
population, $r$~is the rank of the $i$-th individual with respect to its
fitness, and \verb!pop_size! is the population size. 

\item Apply operators to selected parent(s) to obtain possible
replacement(s).  

\item[4a.] Look for an individual
  identical to  possible replacement(s) in the population. If such
  individual(s) exists, no replacement is performed.
 
\item[4b.] Replace old individual if 
\begin{equation}
u ( 0, 1) \leq e^{ \left( F  ( I_{\rm old} ) - F ( I_{\rm new})
\right)/T_t },  
\label{eq:rasa1}
\end{equation}
where $F ( \cdot )$ is the fitness of a given individual, $T_t$ is
the actual temperature and $u( \cdot, \cdot )$ is a random number with 
the uniform distribution on a given interval.

\item Steps 2--4 are performed until the number of successfully
accepted individuals reaches \verb!success_max! or selected number of 
steps reaches \verb!counter_max!.

\item Decrease temperature 
\begin{equation}
T_{t+1} = {\tt T\_mult} T_t .
\end{equation}
If actual temperature $T_{t+1}$ is smaller than $T_{min}$, perform
{\em reannealing} -- i.e. perform step \#1 for one half of the population.

\item Steps 2--6 are repeated until the termination condition is
attained. 
\end{enumerate}

\paragraph{List of operators} The following set of real-valued
operators, proposed in \cite{Michalewicz:1994}, was
implemented. In the sequel, we will denote $L$ and $U$ as vectors of
lower/upper bounds on unknown variables, $u(a,b)$ and $u[a,b]$ as a
real or integer  random variable with the uniform distribution on a
closed interval $\langle a, b \rangle$. Otherwise we use the
same notation as employed in Sections 3.1 and 3.2.

\begin{description}
\item[Uniform mutation:] Let $k = [1, n]$  
\begin{equation}
ch_{ij}( t + 1 ) = \left \{ 
\begin{array}{cl}
 u(L_j, U_j), & {\rm if}\ j=k \\
 ch_{ij} ( t ) , & {\rm otherwise},
\end{array} \right.
\end{equation}
\item[Boundary mutation:] Let $k=u[1,n]$, $p=u(0,1)$ and set:  
\begin{equation}
ch_{ij} ( t+ 1) = \left \{ 
\begin{array}{cl}
 L_j, & {\rm if}\ j=k,p<.5 \\
 U_j, & {\rm if}\ j=k,p\geq.5 \\
 ch_{ij} ( t ), & {\rm otherwise}
\end{array} \right.
\end{equation}

\item[Non-uniform mutation:] Let $k=[1, n]$, $p=u(0,1)$ and set:
\begin{equation}
ch_{ij} (t+1)  = \left \{ 
\begin{array}{cl}
 ch_{ij} ( t ) + ( L_j - ch_{ij} ( t ) ) f, & {\rm if}\ j=k, p<.5 \\
 ch_{ij} ( t ) + ( U_j - ch_{ij} ( t ) ) f ,& {\rm if}\ j=k, p\geq.5 \\
 ch_{ij} ( t ) , & {\rm otherwise}
\end{array} \right.
\end{equation}
where $f = u(0,1)(T_t/T_0)^{\tt b}$ and \verb!b! is the shape
parameter.  

\item[Multi-non-uniform mutation:] Apply non-uniform mutation to all
variables of $CH_i$.   

\item[Simple crossover:] Let $k=[1,n]$  and
set: 
\begin{eqnarray*}
ch_{il} ( t + 1 )  = \left \{ 
\begin{array}{cl}
 ch_{il} ( t ), & {\rm if}\ l<k\\
 ch_{jl} ( t ), & {\rm otherwise}
\end{array} \right. && 
ch_{jl} ( t + 1 ) = \left \{ 
\begin{array}{cl}
 ch_{jl} ( t ), & {\rm if}\ l<k\\
 ch_{il} ( t ), & {\rm otherwise}
\end{array} \right.
\end{eqnarray*}

\item[Simple arithmetic crossover:] Let $k = u[1,n]$, $p=u(0,1)$ and  
set: 
\begin{eqnarray}
ch_{il} ( t + 1 ) & = &  \left \{ 
\begin{array}{cl}
 p ch_{il} ( t )  + (1-p) ch_{jl} ( t ) , & {\rm if}\ l=k \\
   ch_{il} ( t )  , & {\rm otherwise}
\end{array} \right. \\
ch_{jl}( t + 1 )  & = & \left \{ 
\begin{array}{cl}
 p ch_{jl}(t) + (1-p) ch_{il} ( t )  , & {\rm if}\ l=k \\
 ch_{jl} ( t )  , & {\rm otherwise}
\end{array} \right.
\end{eqnarray}

\item[Whole arithmetic crossover:] Simple arithmetic crossover
applied to all variables of $CH_i$ and $CH_j$. 

\item[Heuristic crossover:] Let $p=u(0,1)$, $j=[1,n]$ and $k=[1,n]$
such that $j \neq k$ and set:
\begin{equation}
CH_i ( t + 1 ) = CH_i ( t ) + p ( CH_j ( t )  - CH_k ( t ) ).
\end{equation}
If $CH_i ( t + 1 ) $ is not feasible then a new random number $p$
is generated until the feasibility condition is met or the maximum
number of heuristic crossover applications \verb!num_heu_max! is exceeded.      
\end{description}  

\subsection{Integer augmented simulated annealing.}
%
Before presenting the actual optimization procedure we first introduce
the mapping between representation and search spaces for individual
design variables. Consider ${\bf X}=\{x_1,x_2,\ldots,x_n\}$ as a
vector of $n$ variables, integer or real numbers $x_i$,  defined on a
closed subset of an appropriate domain $D_i\subseteq {\cal N},{\cal
R}$~. Further assume that each variable $x_i$ is represented with some  
required precision $p_i$, defined as the smallest unit the number
$x_i$ can attain. Then, each variable $x_i$ can be
transformed into an integer number $y_i\in {\cal N}$ as
\begin{equation}\label{eq12}
y_i=\bigg[{x_i}{p_i}^{-1}\bigg] \; ,
\end{equation}
where $\left[{x_i}{p_i}^{-1}\right]$ denotes the integer part of
${x_i}{p_i}^{-1}$. 
An inverse transformation is given by
\begin{equation}\label{eq13}
x_i=y_i\;p_i \; .
\end{equation}
For instance, the real number $314.159$ with precision $p_i=0.001$
is transformed into the integer number $314159$. An important aspect
of this methodology is that the encoded number $y_i$ can be treated
either as a binary string using bit-based operations or as a vector of
integer numbers.

Integer augmented simulated annealing method is based on the same
ideas as the previously mentioned RASA algorithm. This procedure
effectively exploits the essentials of GAs (a population of
chromosomes, rather then a single point in space is optimized)
together with the basic concept of simulated annealing method guiding
the search towards minimal energy states. To avoid well-known problems
with binary coding the integer coding was
used. Together with new operators such as differential crossover and a
new mutation operator encouraging results were obtained. 

The description of the algorithm does not substantially differ from the
RASA algorithm, but for the sake of completeness all steps are briefly
reviewed here. 

\begin{enumerate}
\item Initial population consisting of $\verb!OldSize!$ individuals is
created randomly and fitnesses are assigned to each
individual. Starting and ending temperatures $\verb!T_min!$ and
$\verb!T_max!$ are set by the user. 

\item If a real random number $p=u(0,1)$ is smaller than 
$\verb!CrossoverProb!$ the crossover is used, otherwise
the mutation is applied. This step is repeated until the number
$\verb!NewSize!$ of new solutions is obtained.

\item For each individual in a ``new'' population one ``parent'' from
``old'' part is selected. The ``old'' solution is replaced if
\begin{equation}
u ( 0, 1 ) \leq \frac{1}{ 1 + e^{ \left( F  ( I_{\rm old} ) - F ( I_{\rm new})
\right)/T_t }} \; ,
\label{eq:iasa:temp}
\end{equation}
where $F ( \cdot )$, $T_t$ and $u( \cdot, \cdot )$ have the same meaning
as in the previous section. Equation (\ref{eq:iasa:temp}) ensures the 50\%
probability of survival when comparing two solutions with the same
fitness. 

\item Steps 2--3 are performed until the number of successfully
accepted individuals reaches $\verb!SuccessMax!$ or the selected number of 
steps reaches $\verb!CounterMax!$.

\item The actual temperature is decreased by
\begin{equation}
T_{t+1} = T_t \left( \frac{ \tt T\_min }{ \tt T\_{max} }
\right)^{\displaystyle \left( \frac{ \tt CounterMax }
{ \tt TminAtCallsRate * MaxCalls } \right)} \; ,
\end{equation}
where $\verb!TminAtCallsRate!$ determines a fraction of the
maximum allowable number of function calls $\verb!MaxCalls!$ in which
the minimum temperature $\verb!T_min!$ will occur. {\em Reannealing}
step is represented here by setting actual temperature $T_{t+1}$
equal to $\verb!T_max!$. 

\item Steps 2--5 are repeated until the termination condition is
attained. 
\end{enumerate}
In connection with the notation and principles mentioned
in previous sections, integer operators within IASA algorithm have
the following form:
\begin{description}

\item [Differential crossover:] This operator is inspired by the
DE. A new individual $CH_j(t)$ is created from three randomly selected
solutions $CH_p(t)$, $CH_q(t)$ and $CH_r(t)$ according to 
\begin{equation}
CH_j(t+1)=CH_p(t)+u(0.0,CR)(CH_q(t)-CH_r(t)).    
\end{equation}
Note that all vectors
$CH_i$ are integer numbers and also that the influence of the
difference on
the right-hand side randomly varies between zero and {\em cross-rate\/}
$CR$. 

\item [Mutation:] Mutation operator is provided by modifying 
each variable in $CH_j(t)$ to 
\begin{equation}
ch_{ij}(t+1)=ch_{ij}(t) + N(0,\frac{| \; ch_{ij}(t)-ch_{pj}(t) \; |}{2}+1) \; ,
\end{equation}
where $N(\cdot,\cdot)$ is a random integer number derived from the
Gauss normal distribution and $ch_{pj}(t)$ is the $j$-th variable of
a randomly selected vector $CH_p(t)$.
\end{description}

\section{Test computations and results}

Each of the methods introduced in the previous section was tested on
all presented optimization problems. The methodology that has
been used for our computations is based on the following criteria:
\begin{itemize}
\item For each problem and each method the computation was run 100
times to avoid an influence of random circumstances.

\item For all cases, the number of successful runs (which can be
traded as a probability of success or the reliability of the method)
is presented. 

\item If the number of successful runs is non-zero, the average number
of fitness calls of all successful runs is also presented. 

\end{itemize}

Further details of individual function settings and methodology
for results evaluation can be found in the next subsections.

\subsection{Results for the Chebychev problem}
\begin{figure}[h]
\includegraphics[width=14cm]{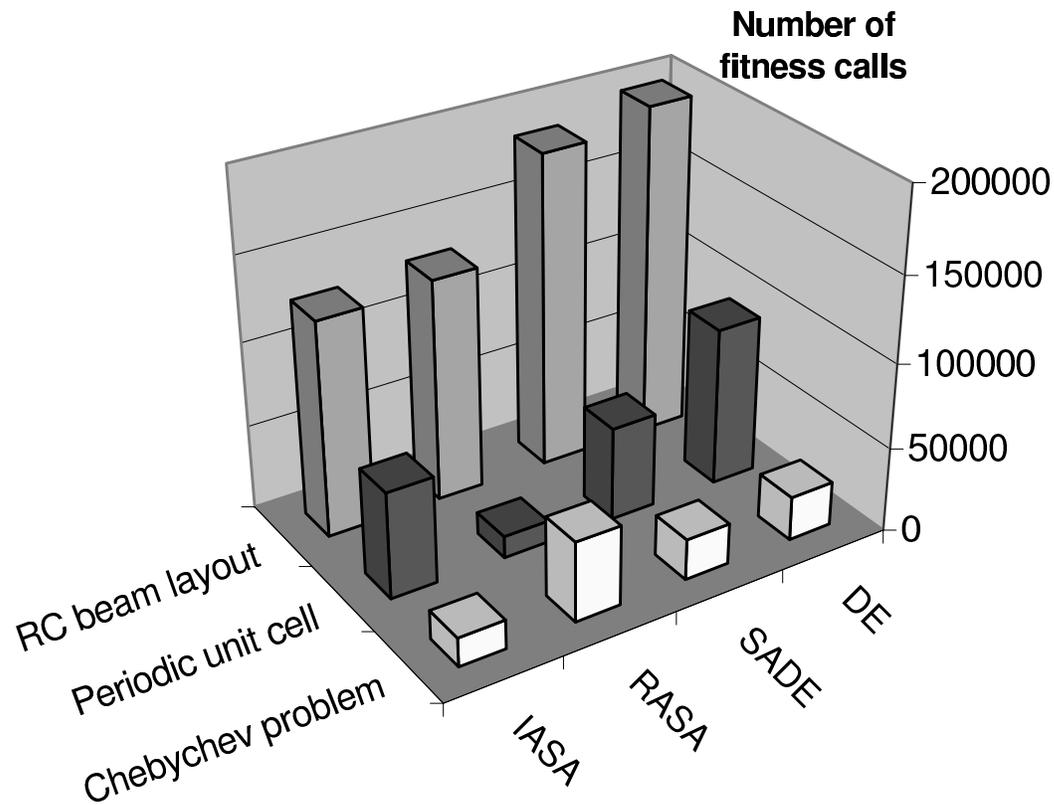}
\caption{A comparison of results for Chebychev polynomial,
  reinforced concrete beam layout and periodic unit cell problems.} 
\label{compare-results}
\end{figure}

The computations were performed for the Chebychev problem with a
degree of the polynomial expression $n=8$ (the T8 problem), which
corresponds to the dimension of the problem $9$. The computation was
terminated if the algorithm reached a value of the objective function
smaller then $10^{-5}$ or the number of function evaluations exceeded
$100,000$. Upper bounds on individual coefficients were set to
$512$, while lower bounds were equal to $-512$. The results of
individual algorithms are shown in Table~\ref{Cheb} and
Figure~\ref{compare-results}.  
\begin{table}[h]
\begin{center}
\begin{tabular}{|l|r|r|r|r|}
\hline
{\bf Method} & {\bf IASA} & {\bf RASA} & {\bf DE} & {\bf SADE} \\
\hline
Successful runs                 & 100   & 100   & 100   & 100 \\
Average number of fitness calls & 10342 & 47151 & 25910 & 24016 \\
\hline
\end{tabular}
\caption{Results for the Chebychev polynomial problem.}
\label{Cheb}
\end{center}
\end{table}

\subsection{Results for the {\em type 0} trial function}
\begin{figure}[h]
\begin{center}
\hspace*{-15mm}
\includegraphics[angle = -90, width = 14cm]{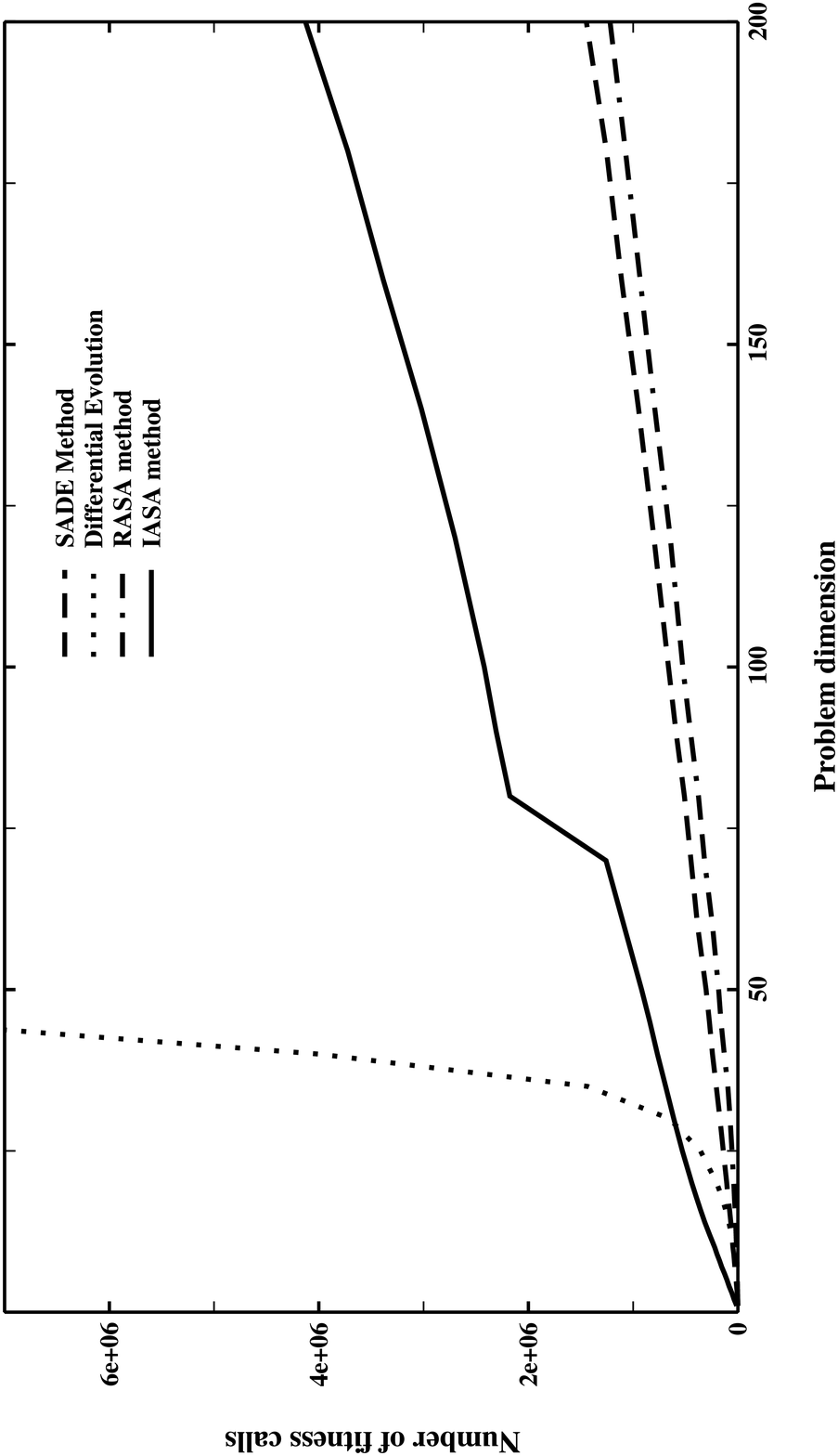}
\\
\bigskip
\caption{A comparison of results for the {\em type 0} function.}
\label{type0-results}
\end{center}
\end{figure}

%
Test computations for the {\em type 0} problem were performed for a 
wide set of problem dimensions, ranging from $1$ to $200$. The upper
bound on each variable was set to $400$, while the lower bound value 
was $-400$. For each run, the position of the extreme was randomly
generated within these bounds and the height of the peak $y_0$ was
generated from the range $0$--$50$. The parameter $r_0$ was set to
$1$. The computation was terminated when the value of the objective 
function was found with a precision greater than $10^{-3}$. The
results are given in the form of the growth of computational
complexity with respect to the problem dimension. For each dimension,
the computation was run 100 times and the average number of fitness
calls was recorded (see~Fig.~\ref{type0-results} and
Table~\ref{t:type0}). 
\begin{table}[h]
\begin{center}
\begin{tabular}{|c|r|r|r|r|}
\hline
{\bf Problem dimension} & {\bf IASA} & {\bf RASA} & {\bf DE} & {\bf SADE} \\
\hline
10  & 246,120 & 13,113    & 39,340  & 46,956  \\
30  & 611,760 & 74,375    & 653,600 & 171,539 \\
50  & 926,100 & 183,882   & N/A     & 304,327 \\
100 & 2,284,590 & 526,492   & N/A     & 663,084 \\
140 & 3,192,800 & 793,036   & N/A     & 948,197 \\
200 & 4,184,200 & 1,220,513 & N/A     & 1,446,540 \\
\hline
\end{tabular}
\caption{Average number of fitness calls for the type-$0$ function}
\label{t:type0}
\end{center}
\end{table}

\subsection{Results for the reinforced concrete beam layout problem}
%
The basic parameters subjected to optimization were the beam width $b$,
which was assumed to take discrete values between $0.15$~m and
$0.45$~m with the step $0.025$~m and the beam height $h$ ranging from
$0.15$~m to $0.85$~m with the step $0.025$~m. For each of the three parts
of a beam, the diameter and the number of longitudinal reinforcing
bars located at the bottom and the top of a beam, spacing
and the diameter of stirrups and the length of the corresponding part were 
optimized. Lower bounds were selected for the sake of structural
requirements; solutions exceeding upper bounds are considered to be
irrelevant for the studied examples. However, from the optimization
point of view, bounds can be easily adjusted to any reasonable
value.The number of longitudinal bars was restricted to the range
$0$--$15$, the spacing of stirrups was assumed to vary from $0.05$~m to
$0.40$~m with the $0.025$~m step. The profiles of longitudinal bars
were drawn from the list of $16$ entries while for the stirrups, only $4$
diameters were considered. This finally results in $18$ independent
variables. Note that the maximal number of longitudinal bars presents
only the upper bound on the searched variable; the specific
restrictions given by Codes of Practice are directly
incorporated in the objective function. For more details
see~\cite{Matej, Honza}. The computation was terminated if an
individual with price smaller than $573.5$~CZK was found or the number
of objective function calls exceeded $1,000,000$.  Table~\ref{mr.beam}
stores the obtained results of different optimization algorithms, see
also Figure~\ref{compare-results}. 
\begin{table}[h]
\begin{center}
\begin{tabular}{|l|r|r|r|r|}
\hline
{\bf Method} & {\bf IASA} & {\bf RASA} & {\bf DE} & {\bf SADE} \\
\hline
Successful runs                 & 100    & 100    & 100    & 100 \\
Average number of fitness calls & 108732 & 131495 & 196451 & 185819 \\
\hline
\end{tabular}
\caption{Results for the reinforced concrete beam layout}
\label{mr.beam}
\end{center}
\end{table}

\subsection{Results for the periodic unit cell problem}
Test computations for the periodic unit cell construction were
performed for the \mbox{$10$-fiber} unit cell (i.e. the dimension of the
problem was $20$). The computation was terminated if algorithm returned
value smaller than $ 6 \times 10^{-5}$ or a number of function calls  
exceeded $400, 000$. Variables were constrained to the box $ 0 \leq x_i 
\leq H_1 \approx 25.8 $ (see Section~(2.4)) . The required numbers of 
function are stored in Table~\ref{puc}~and displayed in
Figure~\ref{compare-results}. 
\begin{table}[h]
\begin{center}
\begin{tabular}{|l|r|r|r|r|}
\hline
{\bf Method} & {\bf IASA} & {\bf RASA} & {\bf DE} & {\bf SADE} \\
\hline
Successful runs                 & 100   & 100   & 100   & 100   \\
Average number of fitness calls & 13641 & 12919 & 93464 & 55262 \\
\hline
\end{tabular}
\caption{Results for the periodic unit cell problem}
\label{puc}
\end{center}
\end{table}

\section{Conclusions}
%
\paragraph{Differential Evolution.}
%
The Differential Evolution algorithm showed to be very efficient and 
robust for moderate-sized problems, but its performance for higher
dimensions deteriorated. Moreover, the small number of
parameters is another advantage of this method. However, the results
suggest that the absence of mutation-type operator(s) is a weak point
of the algorithm.

\paragraph{Simplified Atavistic Differential Evolution.} 
%
The SADE algorithm was able to solve all problems of our test set
with a high reliability and speed. Although it needed larger
number of function calls than other methods (see Table~\ref{Order}),
the differences are only marginal and do not present any serious
disadvantage. Another attractive feature of this method is 
relatively small number of parameters. 
\begin{table}[h]
\begin{center}
\begin{tabular}{|l|c|c|c|c|}
\hline
{\bf Method} & {\bf IASA} & {\bf RASA} & {\bf DE} & {\bf SADE} \\
\hline
{\bf Chebychev problem}    & 1 & 4 & 3 & 2 \\
{\bf Type 0 test function} & 3 & 1 & 4$^\ast$ & 2 \\
{\bf Concrete beam layout} & 1 & 2 & 4 & 3 \\
{\bf Periodic unit cell}   & 3 & 1 & 4 & 2 \\
\hline
{$\bf \Sigma$}             & 8 & 8 & 14 & 9  \\
\hline
\end{tabular}
\caption{Overall performance of methods. ($^\ast$ : Not successful for 
all runs)}
\label{Order}
\end{center}
\end{table}

\paragraph{Real-valued Augmented Simulated Annealing.} 
%
The RASA algorithm was successful for all presented problems; the
average number of function calls was comparable to the other methods. The
obvious disadvantage of this algorithm is a large number of
parameters, which can results in a tedious tuning procedure. On the
other hand, as follows from the Appendix, only two types of parameter
settings were necessary~--~one for the continuous and one for discrete
functions.    

\paragraph{Integer Augmented Simulated Annealing.}
%
The IASA algorithm was the most successful and fastest method on
problems with small dimensions. But on the problems with larger
dimensions and with a higher number of local minima, the algorithm
suffers from premature convergence and limited precision due to
integer coding of variables. In addition, initial tuning of individual  
presents another drawback of this method.

Summary results are given in Table~\ref{Order} to quantify the
overall performance of individual methods. Each of the method is
ranked  primarily with respect to its success rate and secondary with
respect to the average number of fitness calls. The sum then reveals
the overall performance of the method.

\paragraph{Final comments.} According to our opinion, several
interesting conclusions and suggestions can be made from the
presented results. Each of them is discussed in more detail.
\begin{itemize}

\item The performance and robustness of SADE method was
distinguishly better than for DE algorithm. This supports an important
role of a mutation operator(s) in the optimization process. 

\item Although algorithms were developed independently, all  use
some form of differential operator. This shows the remarkable
performance of this operator for both real-valued and discrete
optimization problems. 

\item The most successful methods, SADE and RASA algorithms, both
employ~a variant of ``local mutation''. This operator seems to be
extremely important for higher-dimensional {\em type-0} functions,
where these methods clearly outperform the others.  

\item Slightly better results of RASA method can be most probably 
attributed to the reannealing/restarting phase of the algorithm (a
trivial but efficient tool for dealing with local minima) and to the
search for an identical individual. The procedure for local minima
assessment was implemented to SADE method (see \cite{Ondra2, Ondra-WWW}
for results), incorporation into IASA algorithm is under development. 

\item When comparing methods based on the discrete coding of variables
with real-encoded ones it becomes clear that for continuous functions
the methods with the real coding perform better. Nevertheless, after
implementing new features, like those mentioned before, the performance
is expected to be similar. On the other hand, the advantage of IASA
algorithm is the possibility of its use for discrete combinatorial
problems like the Traveling salesman problem.  

\end{itemize}
Therefore, from the practical point of view, the SADE method seems
to be the most flexible alternative due to its simplicity and small
number of parameters.

\paragraph{Acknowledgement.} We would like to thank an anonymous
referee for his careful revision and comments that helped us to
substantially improve the quality of the paper.The financial support
for this work was provided by the Ministry of Education, projects
No.~MSM~210000003 and MSM~210000015 and by GA\v{C}R grant
103/97/K003. 

\section*{Appendix}
See Tables \ref{t:DE}--\ref{t:IASA}.
\begin{table}[!h]
\begin{center}
\begin{tabular}{|l|l|l|l|l|}
\hline 
\bf Parameter   & \bf Chebychev, Type 0 & \bf Beam & \bf PUC  \\
\hline
\verb!pop_size! & $10\times{\tt dim}$ & $11\times{\tt dim}$ &
$10\times{\tt dim}$ \\ 
$F_1 = F_2 $ & $0.85$ & $0.85$ & $0.75$ \\
$CR$         & $1$    & $0.1$  & $1$    \\
\hline 
\end{tabular}
\end{center}
\caption{Parameter settings for DE}
\label{t:DE}
\end{table}
\begin{table}[!h]
\begin{center}
\begin{tabular}{|l|l|l|l|l|}
\hline 
\bf Parameter & \bf Chebychev & \bf Type 0 & \bf Beam & \bf PUC  \\
\hline 
\verb!pop_size!     & $10\times${\tt dim} & $25\times${\tt dim} & 
$10\times${\tt dim} & $10\times${\tt dim} \\
$CR$                & $0.44$ & $0.1$  & $0.3$  & $0.2$ \\
{\em radioactivity} & $0$    & $0.05$ & $0.05$ & $0.3$ \\
$MR$                & $0.5$  & $0.5$  & $0.5$  & $0.5$ \\
\hline 
\end{tabular}
\end{center}
\caption{Parameter settings for SADE}
\end{table}
\begin{table}[!h]
\begin{center}
\begin{tabular}{|l|l|l|}
\hline 
\bf Parameter & \bf Beam & \bf Others \\
\hline 
\verb!pop_size!  & $64$     & $32$   \\
\verb!q!         & $0.04$   & $0.04$ \\
\verb!p_uni_mut! & $0.525$ & $0.05$ \\
\verb!p_bnd_mut! & $0.125$ & $0.05$ \\
\verb!p_nun_mut! & $0.125$ & $0.05$ \\
\verb!p_mnu_mut! & $0.125$ & $0.05$ \\
\verb!p_smp_crs! & $0.025$ & $0.15$ \\
\verb!p_sar_crs! & $0.025$ & $0.15$ \\
\verb!p_war_crs! & $0.025$ & $0.15$ \\
\verb!p_heu_crs! & $0.025$ & $0.35$ \\
\verb!b!         & $0.25$  & $2.0$  \\
\verb!T_frac!     & $10^{-2}$ & $10^{-10}$ \\
\verb!T_frac_min! & $10^{-4}$ & $10^{-14}$ \\
\verb!T_mult!     & $0.9$     & $0.9$ \\
\verb!num_success_max! & $10\times$\verb!pop_size! & $10\times$\verb!pop_size! \\
\verb!num_counter_max! & $50\times$\verb!pop_size! & $50\times$\verb!pop_size! \\
\verb!num_heu_max!     & $20$   & $20$  \\
\verb!precision! (step 4a) & see Section 4.3 & $ 10^{-4}$ \\
\hline 
\end{tabular}
\end{center}
\caption{Parameter settings for RASA}
\end{table}
\begin{table}[!h]
\begin{center}
\begin{tabular}{|l|l|l|l|l|}
\hline 
\bf Parameter & \bf Chebychev & \bf Type 0 & \bf Beam & \bf PUC  \\
\hline 
\verb!OldSize!         & $80$     & $900$      & $180$      & $200$\\
\verb!NewSize!         & $5$      & $600$      & $250$      & $100$ \\
\verb!T_max!           & $10^{-5}$ & $10^{-5}$  & $10^{-4}$ & $10^{-1}$ \\
\verb!T_min!           & $10^{-7}$ & $10^{-10}$ & $10^{-5}$ & $10^{-5}$ \\
\verb!SuccessMax!      & $1000$    & $1000$     & $1000$    & $1000$ \\
\verb!CounterMax!      & $5000$    & $5000$     & $5000$    & $5000$ \\
\verb!TminAtCallsRate! & $19\%$    & $100\%$    & $25\%$    & $20\%$ \\
\verb!CrossoverProb!   & $97\%$    & $92\%$     & $60\%$    & $90\%$ \\
$CR$                   & $0.5$     & $0.6$      & $1.3$     & $1.0$ \\
\hline
\end{tabular}
\end{center}
\caption{Parameter settings for IASA}
\label{t:IASA}
\end{table}

\end{document}